\documentclass[letterpaper, 10 pt, conference]{ieeeconf}  

\IEEEoverridecommandlockouts                              

\usepackage{xcolor}
\usepackage{url}
\usepackage{algpseudocode}
\usepackage{algorithm}
\usepackage{amsmath}
\usepackage{graphicx}  
\usepackage{tikz}
\usetikzlibrary{arrows.meta, calc, positioning}
\usepackage{booktabs}
\usepackage{graphicx}
\usepackage{xcolor}
\usepackage[most]{tcolorbox}

\usepackage{cite}
\newcommand{\revise}[1]{\textcolor{black}{\textbf{}{#1}}}

\title{\LARGE \bf
Dialogue based Interactive Explanations for Safety Decisions in Human Robot Collaboration
}

\author{
Yifan Xu$^{1\dagger}$,
Xiao Zhan$^{2\dagger}$,
Akilu Yunusa Kaltungo$^{3}$,
Ming Shan Ng$^{4}$,\\
Tsukasa Ishizawa$^{5}$,
Kota Fujimoto$^{6}$,
Clara Cheung$^{1*}$
\thanks{*Corresponding: ({\tt\small clara.cheung@manchester.ac.uk}).}%
\thanks{$\dagger$ Both contributed equally to this work.}%
\thanks{$^{1}$Department of Civil Engineering and Management, Faculty of Science and Engineering, The University of Manchester, Manchester, United Kingdom
}
\thanks{$^{2}$VRAIN, Universitat Politècnica de València, Valencia, Spain \& Department of Engineering, University of Cambridge, Cambridge, United Kingdom
}
\thanks{$^{3}$Department of Mechanical and Aerospace Engineering, Faculty of Science and Engineering, The University of Manchester, Manchester, United Kingdom
}%
\thanks{$^{4}$Center for the Possible Futures, Kyoto Institute of Technology, Kyoto, Japan
}%
\thanks{$^{5}$Institute of Industrial Science, The University of Tokyo, Japan
}%
\thanks{$^{6}$Graduate School of Frontier Sciences,
The University of Tokyo, Japan
}%
}

\begin{document}

\maketitle
\thispagestyle{empty}
\pagestyle{empty}

\begin{abstract}

As robots increasingly operate in shared, safety-critical environments, acting safely is no longer sufficient robots must also make their safety decisions intelligible to human collaborators. In human robot collaboration (HRC), behaviours such as stopping or switching modes are often triggered by internal safety constraints that remain opaque to nearby workers. We present a dialogue based framework for interactive explanation of safety decisions in HRC. The approach tightly couples explanation with constraint-based safety evaluation, grounding dialogue in the same state and constraint representations that govern behaviour selection. Explanations are derived directly from the recorded decision trace, enabling users to pose causal (“Why?”), contrastive (“Why not?”), and counterfactual (“What if?”) queries about safety interventions. Counterfactual reasoning is evaluated in a bounded manner under fixed, certified safety parameters, ensuring that interactive exploration does not relax operational guarantees. We instantiate the framework in a construction robotics scenario and provide a structured operational trace illustrating how constraint aware dialogue clarifies safety interventions and supports coordinated task recovery. By treating explanation as an operational interface to safety control, this work advances a design perspective for interactive, safety aware autonomy in HRC.


\end{abstract}


\section{Introduction}


As autonomous robots increasingly move into shared, safety-critical environments, they are required not only to act safely but also to make their safety decisions understandable to human collaborators~\cite{alonso2018system,setchi2020explainable}. In domains such as construction, logistics, and infrastructure maintenance, robots continuously evaluate proximity, visibility, task priority, and environmental uncertainty to determine whether to proceed, slow down, stop, or switch control modes~\cite{li2024safe}. While these safety interventions are often technically sound, their underlying reasoning is rarely accessible to human partners in real time. Current safety communication mechanisms in human–robot interaction primarily rely on indicator lights, warning sounds, or short textual status messages~\cite{orthmann2023sounding,tang2019development,10.1145/3171221.3171249,san2025mixed,cini2021relevance}. These signals indicate that a constraint has been triggered, but seldom explain \emph{why} a decision was taken, \emph{why not} an alternative was allowed, or \emph{under what conditions} the task could resume. Prior work in explainable robotics has explored transparency, post-hoc rationalization, and confidence reporting~\cite{yadollahi2024explainability,li2024safe}. However, explanation is often treated as a separate interface layer rather than as part of the safety control process itself.

Research in explainable AI suggests that human explanations are inherently contrastive and counterfactual in nature: people do not merely ask “What happened?”, but rather “Why this instead of that?”~\cite{miller2019explanation, hayes2017improving}. Similarly, counterfactual reasoning has been identified as a central mechanism for making automated decisions intelligible~\cite{wachter2017counterfactual}. In planning and decision-support systems, structured “Why?” and “Why-not?” queries have been used to reconcile differences between system and user models~\cite{westphal2023decision, chakraborti2017plan,sreedharan2019can}. Yet these approaches are typically developed for symbolic planning contexts, where reasoning traces are discrete and static. 

Safety critical HRC introduces fundamentally different conditions. Robot behaviour is governed not only by symbolic rules, but by continuously evaluated safety constraints grounded in physical state, sensing uncertainty, and certified operational limits. Classical work on safe physical human–robot interaction emphasises the importance of maintaining explicit safety envelopes during collaboration~\cite{de2006collision}. When a robot stops due to proximity, occlusion, or uncertainty, the cause is not a failed logical proof but the activation of one or more safety constraints. At the same time, shared task performance depends on aligned mental models and calibrated trust between human and robot~\cite{hancock2011meta}. If safety interventions remain opaque, human collaborators may misinterpret robot intent, over-trust, or under-trust the system. Thus, explanation in safety-critical HRC must serve not merely as transparency, but as a mechanism for maintaining shared situational awareness during task interruption and recovery.

In this paper, we present a dialogue based framework for safety grounded explanation in HRC. The central idea is to tightly couple dialogue generation with constraint based safety evaluation. At each time step, the robot maintains a structured safety state and records the active constraints that determine behaviour selection. Dialogue responses are derived directly from this decision trace: causal queries retrieve the triggering constraint; contrastive queries identify which safety parameter prevents an alternative action; counterfactual queries construct a hypothetical state and re-evaluate feasibility under the unchanged safety envelope. Rather than proposing explanation as a post-hoc narrative layer, this work treats explanation as an operational interface to the robot’s safety logic. By grounding dialogue in the same mechanisms that enforce behavioural constraints, the framework aims to support human understanding while preserving formally defined safety limits.


\begin{table*}[t]
\centering
\caption{Safety grounded dialogue query types and their corresponding explanation mechanisms derived from the decision trace}
\label{tab:safety_queries}
\renewcommand{\arraystretch}{1.0}
\begin{tabular}{p{4cm} p{12cm}}
\toprule
\textbf{User Question} & \textbf{Safety Grounded Explanation Mechanism} \\
\midrule

Why did you stop?
& The robot points to the safety constraint in $C_t$ 
that directly triggered the current behavior $u_t$ 
(e.g., the worker entered the 1.5\,m safety zone). \\[4pt]

Why didn’t you continue lifting?
& The robot considers the alternative behavior and explains 
which safety parameter in $P$ would be violated 
if that behavior were executed. \\[4pt]

What if I step back? 
& The robot evaluates the proposed change by constructing 
a hypothetical safety state $S_t'$, recomputing active constraints, 
and explaining whether a different behavior becomes feasible 
or remains blocked. \\

\bottomrule
\end{tabular}
\end{table*}


\section{Dialogue Based Safety Explanation Framework}
\label{sec:framework}

This work builds on our prior dialogue based explanation framework ~\cite{xu2024explanation}, which conceptualized explanation as a structured, multi-turn interaction grounded in explicit reasoning traces. In that setting, users interrogate the system through regulated dialogue moves (e.g., ``Why?'' and ``Why not?'') over symbolic inference structures, enabling targeted clarification of specific reasoning steps under disagreement or uncertainty. Safety critical HRC, however, introduces fundamentally different conditions. \revise{Robot behaviour is governed not only by symbolic reasoning but by continuously evaluated safety conditions coupling the human state, robot state, and environment~\cite{tonola2025reactive,adesiji2026safety}}. These evaluations are driven by real-time sensing updates, uncertainty, and time-sensitive control decisions. As a result, when a robot stops, slows down, or switches modes, the underlying cause is typically the activation of one or more safety constraints rather than the outcome of a static proof.

We therefore reinterpret dialogue based explanation in safety critical HRC as an operational interface to the robot's safety controller. The key design requirement is groundedness: explanations must be derived from the same state variables, constraints, and parameters that govern behaviour selection. This grounding enables the robot to answer the types of questions collaborators naturally ask during interruptions \emph{Why did you stop? Why not continue? What if I move?} in a way that is both interpretable and consistent with certified safety limits. Figure~\ref{fig:dialogue_pipeline} provides an overview of the resulting pipeline.



\subsection{Safety Grounded Decision State}

In safety critical HRC robot behaviour is governed by continuously evaluated safety constraints rather than by a purely task-driven objective. Before executing any action, the system must verify that the current situation satisfies certified safety requirements. To make this process explicit and explainable, we formalise the information available to the safety controller at each time step.

At time $t$, the robot maintains a structured safety state:
\begin{equation}
S_t = \langle H_t, R_t, E_t, P \rangle,
\end{equation}
where:

\begin{itemize}
    \item $H_t$ describes the human-related state 
    (e.g., worker position, motion direction, role).
    \item $R_t$ describes the robot’s internal condition 
    (e.g., pose, velocity, load status).
    \item $E_t$ captures relevant environmental context 
    (e.g., occlusions, nearby moving equipment).
    \item $P$ represents the active safety parameters 
    (e.g., minimum separation distance, visibility thresholds).
\end{itemize}

The tuple $S_t$ summarises all information required to evaluate whether continued task execution is safe under the current operational envelope. It captures the information needed to answer a fundamental question: \emph{Is it safe to continue the task?} Based on $S_t$, a nominal task policy proposes an action $u_t$ (e.g., \textit{continue}, \textit{slow down}, \textit{stop}, or \textit{switch mode}). Before execution, however, the candidate action is evaluated against the safety parameters encoded in $P$. If any safety constraint is violated for example, if a worker enters a protected zone the safety controller overrides the nominal task plan and enforces a safer alternative.


Safety is enforced through constraint functions that evaluate whether specific safety conditions are violated. Let $C_t$ denote the set of active constraints at time $t$. If $C_t = \emptyset$, the robot executes the nominal task action. If one or more constraints are triggered, the safety controller selects a safe alternative behaviour according to a predefined safety-priority structure. To support interactive explanation, the system records a structured decision trace:
\begin{equation}
D_t = \{S_t, u_t, C_t\},
\end{equation}

where $C_t$ denotes the safety constraints that were active when $u_t$ was determined. For example, $C_t$ may indicate that the robot stopped because the worker’s distance fell below 1.5\,m or because visibility dropped below an acceptable threshold. Explanations refer directly to these recorded constraints, ensuring that they reflect the same safety reasoning that governed the robot’s behaviour.

\subsection{Dialogue Mechanism and Safety Grounded Reasoning}
Explanations are provided through short, task oriented dialogue between the human collaborator and the robot. Rather than generating a single static justification, the system supports multi-turn interaction that remains structurally coupled to the safety controller. This coupling is particularly important in construction environments, where safety interventions occur frequently and must be understood rapidly to avoid unnecessary disruption of workflow. \revise{The robot maintains a lightweight dialogue memory $M_t$ that tracks shared context (e.g., which worker, obstacle, or safety zone is being discussed). The memory supports clarification, avoids redundant explanations, and enables refinement across turns.} \revise{In practice, safety related explanatory questions typically fall into 
three common types: \emph{Why?}, \emph{Why not?}, and \emph{What if?}} each corresponding to a distinct reasoning operation grounded in the safety controller. Table~\ref{tab:safety_queries} summarises these categories. Figure~\ref{fig:dialogue_pipeline} illustrates how these query types are operationalised within the safety grounded explanation pipeline. 
\begin{figure}[t]
\centering
\resizebox{\columnwidth}{!}{%
\begin{tikzpicture}[
  font=\small,
  node distance=5mm and 6mm,
  box/.style={draw, rounded corners, align=center, inner sep=3pt, minimum height=8mm, text width=40mm},
  sbox/.style={draw, rounded corners, align=center, inner sep=3pt, minimum height=8mm, text width=26mm},
  arrow/.style={-Latex, line width=0.6pt}
]

\node[box] (Q) {User Query $Q$\\\footnotesize(e.g., ``Why did you stop?'')};
\node[box, below=of Q] (Route) {Classify query\\\footnotesize Why / Why-not / What-if};
\node[box, below=of Route] (State) {Decision state\\$D_t=\{S_t,u_t,C_t\}$\\\vspace{1pt}\footnotesize $C_t$ from $(S_t,P)$};

\node[sbox, below left=6mm and 6mm of State] (Why) {Why\\\footnotesize use trigger constraint in $C_t$};
\node[sbox, below=6mm of State] (WhyNot) {Why-not\\\footnotesize find violated\\constraint(s)};
\node[sbox, below right=6mm and 6mm of State] (WhatIf) {What-if\\\footnotesize form $S_t'$\\recompute $C_t'$};

\node[box, below=25mm of State] (E) {Explanation\\$E$};
\node[box, below=of E] (Bound) {Certified safety\\envelope $P$};

\draw[arrow] (Q) -- (Route);
\draw[arrow] (Route) -- (State);

\draw[arrow] (State) -- (Why);
\draw[arrow] (State) -- (WhyNot);
\draw[arrow] (State) -- (WhatIf);

\draw[arrow] (Why) -- (E);
\draw[arrow] (WhyNot) -- (E);
\draw[arrow] (WhatIf) -- (E);

\draw[arrow] (E) -- (Bound);

\end{tikzpicture}%
}
\caption{Safety grounded dialogue explanation pipeline. Queries (e.g., ``Why did you stop?'', ``Why not continue?'', ``What if I move closer?'') are classified into \emph{Why}, \emph{Why-not}, or \emph{What-if}. Explanations are grounded in the recorded decision state $D_t=\{S_t,u_t,C_t\}$. Counterfactual queries form a hypothetical state $S_t'$ and re-evaluate constraints ($C_t'$) within certified safety limits $P$.}
\label{fig:dialogue_pipeline}
\end{figure}

Explanation depends on both the recorded decision state and the evolving dialogue context:
\begin{equation}
E = Explain(D_t, Q, M_t),
\end{equation}
where $Q$ denotes the user query and $M_t$ captures the dialogue memory. 
Responses are derived directly from the decision trace 
$D_t = \{S_t, u_t, C_t\}$, ensuring consistency with the underlying safety controller.  For counterfactual queries, the system constructs a hypothetical state $S'_t$ reflecting the proposed modification and re-evaluates the safety 
constraints under the same certified parameter set $P$. 
An alternative action is considered feasible only if all safety 
constraints remain satisfied, ensuring that counterfactual reasoning 
remains strictly bounded within verified safety envelopes.








\section{Construction Case Study}
\begin{figure}[t]
\centering
\includegraphics[width=\linewidth]{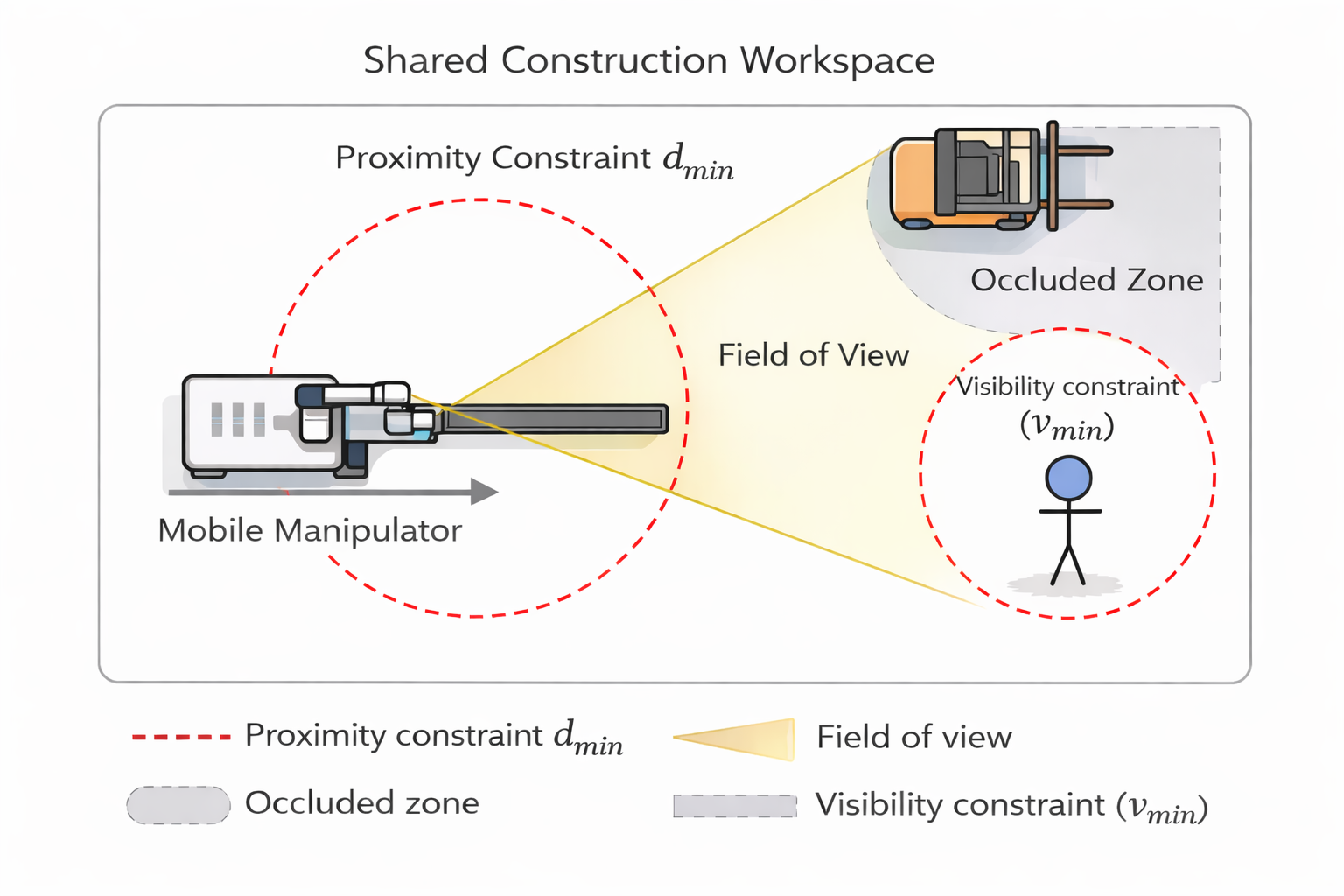}
\caption{
Construction environment schematic. 
A mobile manipulator transports a steel beam in a shared workspace 
with a human worker and a forklift. 
Safety envelopes include a proximity constraint $d_{\min}$ 
(dashed circle around the robot) and a visibility constraint 
defined by a minimum confidence level $v_{\min}$ 
(illustrated within the worker region). 
The forklift creates an occluded zone that reduces sensing reliability, 
potentially activating the visibility constraint.
}
\label{fig:construction}
\end{figure}
We demonstrate the framework through a representative construction scenario in which a mobile manipulator transports steel beams across a partially obstructed shared workspace (Fig.~\ref{fig:construction}). 
The environment contains human workers, dynamic equipment such as forklifts, and structural occlusions that intermittently affect sensing reliability. Following the safety grounded representation introduced earlier, the robot maintains a state $S_t = \langle H_t, R_t, E_t, P \rangle$. In the depicted scenario, a forklift crosses the robot’s field of view, 
reducing visibility confidence below the required minimum level $v_{\min}$. The corresponding constraint becomes active,
\[
C_t = \{\texttt{visibility} < v_{\min}\},
\]
and the robot selects the behaviour
\[
u_t = \texttt{pause}.
\]

\revise{Instead of a generic ``Safety Stop" indicator, the dialogue framework enables structured interaction:}
\begin{figure}[h!]
\centering
\scalebox{0.92}{
\begin{tcolorbox}[colback=gray!5!white,
colframe=black,
title=Dialogue-Based Safety Explanation]
\textbf{Human:} Why did you stop?\\
\textbf{Robot:} My visibility confidence dropped below the required threshold due to occlusion in Zone C.\\
\textbf{Human:} Was it the forklift near the column?\\
\textbf{Robot:} Yes. The forklift obstructed my field of view 
and reduced sensing reliability.\\
\textbf{Human:} What if I guide you past it manually?\\
\textbf{Robot:} If you remain within one meter on my right side, I can switch to manual-follow mode while maintaining safe clearance.\\
\textbf{Human:} Do it. Let’s go.\\
\textbf{Robot:} Switching to manual-follow.
Please remain within the guidance zone.
\end{tcolorbox}
}
\caption{Dialogue-based safety in the construction scenario.}
\label{fig:construction_dialogue}
\end{figure}





The initial \emph{Why} query retrieves the binding constraint in $C_t$ responsible for the current behaviour. The clarification aligns shared reference without altering the safety state. 
The \emph{What-if} query triggers bounded counterfactual reasoning: the robot constructs a hypothetical state $S_t'$ reflecting the proposed manual-guidance condition and reassesses feasibility under the same safety parameters $P$. Under controlled proximity, the visibility constraint is mitigated, and a new feasible behaviour is selected:
\[
u_t' = \texttt{manual-follow}.
\]

Explanation is therefore not a post-hoc justification, but a control-coupled interaction loop. By exposing constraint boundaries and evaluating safe alternatives in real time, the robot supports negotiated task recovery while preserving formally defined safety guarantees. A similar interaction can occur when visibility constraints become active. 
For example, a forklift may temporarily block the robot's line of sight 
while a worker moves behind stacked materials. In this situation, the 
perception module reports a visibility confidence of $0.52 < v_{\min}=0.6$, 
activating the visibility constraint and triggering a slowdown command. 
When asked “Why did you slow down?”, the robot explains that sensing 
confidence dropped below the safe threshold.



\subsection{Prototype Instantiation}

To make the proposed framework concrete, we construct a lightweight rule-based instantiation of a mobile
manipulator operating in a shared construction workspace. The instantiation operationalises the safety grounded decision model by coupling constraint evaluation with dialogue-based explanation. At each time step, the safety controller evaluates the safety state $S_t$ and determines the active constraint
set $C_t$. Each constraint is defined as a condition over the state variables and maps to a corresponding behaviour $u_t$. Formally, constraints are expressed as
\[
\text{condition}(S_t) \;\Rightarrow\; u_t,
\]
where behaviours are selected only if all safety parameters in $P$ are satisfied. If multiple constraints are active, selection follows a predefined safety priority ordering. Representative safety constraints are summarized in Table~\ref{tab:safety_rules}. 


\renewcommand{\arraystretch}{1.2}
\begin{table}[H]
\centering
\caption{Representative safety constraints in the prototype.}
\label{tab:safety_rules}
\begin{tabular}{p{5cm} p{2.5cm}}
\hline
\textbf{Constraint Condition (over $S_t$)} & \textbf{Selected Behavior} \\
\hline
$d(H_t, R_t) < d_{\min}$ 
& $u_t = \texttt{stop}$ \\

$\texttt{visibility}(E_t) < v_{\min}$ 
& $u_t = \texttt{pause}$ \\

$\texttt{worker\_in\_guidance\_zone}(H_t)$ 
& $u_t = \texttt{manual}$ \\
\hline
\end{tabular}
\end{table}

When constraints become active, the selected behaviour $u_t$ and the corresponding constraint set $C_t$ are recorded as part of the decision state $D_t$. This decision trace serves as the grounding structure for dialogue responses. For causal queries, the dialogue module retrieves the binding constraints from $C_t$. For contrastive queries, it evaluates which safety parameter in $P$ would be violated by an alternative action. For counterfactual queries, the dialogue manager constructs a hypothetical state $S_t'$ reflecting the proposed modification and re-applies the same constraint evaluation under the unchanged safety envelope $P$. A new behaviour $u_t'$ is considered feasible only if the modified state satisfies all certified constraints.


\section{Discussion and Future Work}

This work demonstrates how safety evaluation and explanation can be structurally integrated within a dialogue-based framework for human–robot collaboration. By grounding interaction in the same constraint-based logic that governs behaviour selection, explanation becomes an operational component of safety control rather than a retrospective justification layer. \revise{Because explanations are derived directly from the recorded decision state $D_t$, explanation generation introduces minimal computational overhead relative to the underlying safety controller, making the approach compatible with real-time operation.} In dynamic environments such as construction sites, this coupling enables rapid clarification of safety interventions while preserving certified limits. 
Identifying which constraints are active, and under what state modifications alternative behaviours would become feasible, supports shared situational awareness during task interruption and recovery without relaxing safety guarantees. The current instantiation is rule-based and demonstrated through a structured simulation trace. It does not yet include uncertainty-aware modelling or empirical user evaluation. Future work includes (i) integrating confidence-aware constraint activation to better reflect perceptual uncertainty, (ii) developing user-adaptive explanation strategies based on role and expertise, and (iii) scaling the approach to multi-agent or multi-robot collaboration settings. As autonomous systems enter increasingly risk-sensitive domains, aligning control mechanisms with explanation design may become as important as advances in sensing or planning.


\section{Conclusion}

This paper presented a dialogue-based framework for safety grounded explanation in HRC. By integrating dialogue directly with constraint based safety evaluation, explanation becomes an operational interface to safety control rather than a post-hoc narrative layer. Grounding responses in the recorded decision trace enables causal, contrastive, and bounded counterfactual queries to be resolved using the same logic that governs behaviour selection. Through a construction robotics scenario, we demonstrated how making active constraints explicit and evaluating alternatives within fixed safety limits can clarify safety interventions and support coordinated task recovery without relaxing certified guarantees. While the current instantiation focuses on a structured operational demonstration, future work will examine its impact on trust calibration, shared mental models, and collaborative performance, as well as extensions to uncertainty-aware and multi-agent settings. We view this work as a step toward more tightly coupling control architectures and explanation mechanisms in safety critical autonomy, where interpretability is a core component of collaborative performance.

\addtolength{\textheight}{-12cm}   









\bibliographystyle{IEEEtran} 
\bibliography{reference}

\end{document}